\documentclass[lettersize,journal]{IEEEtran}
\usepackage{amsmath,amsfonts}
\usepackage{algorithmic}
\usepackage{algorithm}
\usepackage{array}
\usepackage[caption=false,font=normalsize,labelfont=sf,textfont=sf]{subfig}
\usepackage{textcomp}
\usepackage{stfloats}
\usepackage{url}
\usepackage{verbatim}
\usepackage{graphicx}
\usepackage{cite}

\usepackage{xcolor}
\usepackage{svg}

\usepackage{caption}
\usepackage{setspace}
\usepackage{tabularx}
\usepackage{makecell}
\usepackage{multirow}
\usepackage{float}

\begin{document}

\title{Bi-directional Loop Closure for Visual SLAM}

\author{Ihtisham Ali, Sari Peltonen, and Atanas Gotchev
        % <-this % stops a space
\thanks{Manuscript received xxx ; revised xxx.}% <-this % stops a space
\thanks{Ihtisham Ali, Sari Peltonen, and Atanas Gotchev are affiliated with Faculty of Information Technology and Communication Sciences, Tampere University, Finland.}
\thanks{Corresponding author: Ihtisham Ali (e-mail: ihtisham.ali@tuni.fi; ihtishamaliktk@gmail.com).}}

% The paper headers
\markboth{Preprint File, April~2022}%
{ \MakeLowercase{}}

\maketitle

\begin{abstract}
A key functional block of visual navigation system for intelligent autonomous vehicles is Loop Closure detection and subsequent relocalisation. State-of-the-Art methods still approach the problem as uni-directional along the direction of the previous motion. As a result, most of the methods fail in the absence of a significantly similar overlap of perspectives. 
In this study, we propose an approach for bi-directional loop closure. This will, for the first time, provide us with the capability to relocalize to a location even when traveling in the opposite direction, thus significantly reducing long-term odometry drift in the absence of a direct loop. 
We present a technique to select training data from large datasets in order to make them usable for the bi-directional problem. The data is used to train and validate two different CNN architectures for loop closure detection and subsequent regression of 6-DOF camera pose between the views in an end-to-end manner. The outcome packs a considerable impact and aids significantly to real-world scenarios that do not offer direct loop closure opportunities.  We provide a rigorous empirical comparison against other established approaches and evaluate our method on both outdoor and indoor data from the FinnForest dataset and PennCOSYVIO dataset. 
\end{abstract}


\section{Introduction}
\IEEEPARstart{I}{nffering} where you are on a map of your local world is a core problem of mobile robotics, navigation, and augmented reality. It refers to the process of recognizing a previously visited place and determining your current pose w.r.t the previous pose from the visual scene representation. This problem is widely known as the lost or kidnapped robot. Loop closure is essential to attain robust navigation in any intelligent transportation system as it aids in significantly reducing the accumulated errors during visual navigation \cite{gao2018ldso}. Traditionally, loop closure is detected in an environment that has been previously viewed from a similar perspective e.g., a vehicle traveling toward the north passes by the same location moving in the same direction. Such a configuration maximizes the chances of place recognition. Both classical methods, such as Binary feature descriptors in conjunction with Bag-of-Words (BoWs), and deep learning-based approaches can be used to tackle this problem, with the latter approach exhibiting better performance after the advent of modern Convolutional Neural Network (CNN) architectures. In this paper, we propose an approach to expand the capability of loop closure detection methods towards bi-directional problems.  Our proposed approach is able to correctly recognize previously visited places and find the relative pose irrespective of the direction of motion of the vehicle. 
Due to the recent success of CNNs in the fields of image classiﬁcation \cite{anwar2017computer,krizhevsky2012imagenet}, semantic segmentation \cite{hong2015decoupled,noh2015learning}, and image retrieval \cite{babenko2014neural,gordo2016deep}, many researchers opt to use them to predict camera pose from visual data \cite{kendall2017geometric,kendall2015posenet}. In some studies, the learned deep feature vectors are directly used for scene recognition following the concepts of image retrieval. However, we show that decoupling the two tasks can result in higher accuracy and better generalization.

Firstly, we introduce a novel automated technique to leverage the use of existing large datasets for training CNNs towards the task of bi-directional loop closure. This is essential since acquiring new data every time requires considerable time and resources. Then, we present two CNN based networks that are assigned to first identify potential candidates for loop closure between the query and database and subsequently regress the pose between the matched candidates. Our tests show that the models exhibit a successful learning pattern, and the feature vectors can be used for scene recognition and pose regression upon slight alterations between the networks. Moreover, the generalization is successfully tested on unseen data thus exhibiting strong comprehension of the visual cues by the model and not just scene memorization. 

The article is organized as follows: In Section 2, we discuss the relevant studies from the literature, their impact, and their limitations. In Section 3, we give out the system overview and formulate the proposed approaches for place recognition and pose regression tasks. Moreover, we also discuss in detail, the data preparation steps involved in this study for the task at hand. Section 4 provides the experimental results and a comprehensive comparison with other well-established methods on two challenging datasets. Finally, Section 5 concludes the article.

\section{Related Works}
%\subsection{Visual Place Recognition}
Traditionally, the place recognition problem has been approached in a similar manner as the Image retrieval problem. In general, a query image, whose location needs to be estimated, is compared against a large geo-tagged database of images from previous visits. Each image is represented as an aggregate of numerous local invariant features. The state-of-the-art method still relies on feature detectors and descriptors such as SIFT \cite{lowe2004distinctive}, ORB \cite{rublee2011orb}, SURF \cite{bay2008speeded}, etc., that are used to extract local information from an image and accumulated into a single feature vector for an entire image using encoded representation through methods such as bag-of-visual-words \cite{philbin2007object}, vector of locally aggregated descriptors (VLAD) \cite{arandjelovic2013all} or Fisher vector \cite{jegou2011aggregating}. Fisher Vector adopts the Gaussian mixture model (GMM) to build a visual word dictionary matching. As a result, Fisher Vector encodes more image information than BoW. and at times outperforms BoW in some computer vision tasks. In contrast, VLAD is a simplification of Fisher Vector and provides a trade-off between performance and computational efficiency. In most cases, VLAD performs similarly to Fisher with better efficiency.  These methods aid in compressing the image representation and subsequent efficient retrieval of a match from the database \cite{jegou2010product}. A popular approach based on the aforementioned concepts is FAB-MAP \cite{cummins2008fab} which learns a generative model for the BoW data. The model observes and learns the co-occurrence of appearance words from common objects that are likely to appear or disappear together thus providing valuable probabilistic information. However, FAB-MAP proves to be computationally expensive due to its complex methodology for image description and matching.

On the other hand, recent studies have shown that the deployment of CNNs results in significant improvement in accuracy and reduction in complexity. The models trained on very large datasets significantly outperform the local descriptors such as SIFT in a variety of applications such as object and scene recognition \cite{memon2020loop}. McManus et al. \cite{mcmanus2014scene} proposed to learn features from image patches and called them scene signatures. These scene signatures were for matching and retrieving scenes under varying appearance changes. However, the approach required a considerably more careful training phase with data of the test environment under all possible environmental conditions. Some studies directly opt for using the intermediate representations which are learned using object recognition dataset and use them for scene identification \cite{chen2014convolutional, sunderhauf2015performance}. Sunderhauf et al. \cite{sunderhauf2015performance} propose the use of features from intermediate layers to form a descriptor for matching. Features from higher layers of a CNN encode semantic information about the place while features from the lower layer encode more descriptive information about the geometry of the scene. The authors experiment with varying combinations of these descriptions and attempt to find the nearest neighbor based on the cosine distance between the feature vectors of the query and the database.

It is noteworthy that all the aforementioned studies targeted uni-directional or traditional loop closure cases. The closest work done to bi-directional loop closure is \cite{arroyo2014bidirectional}. The authors claim to target the problem of bi-directional loop closure in panoramic images. In our humble opinion, the use of panoramic images diminished the complications of the problem by providing roughly similar views to a uni-directional case. The panoramas are captured in an enclosed structural environment with a circular trajectory. As a result, the motion in the reverse direction captures a considerable overlap of the forward motion scenes with some spatial offset in images and only marginal difference in perspectives. This can also be observed from the illustrations in the study which exhibit only spatial changes in the scene. Moreover, the study also reports that traditional methods such as FAB-MAP fail to close the loop in practice even in these panoramic images.
To the best of our knowledge, this is the only study that targets place recognition and loop closure while moving in the opposite direction and high perspective change.
In our work, we propose the use of a CNN topped with a neural network implementation of VLAD known as NetVLAD \cite{arandjelovic2016netvlad}.  The model learns to recognize places in an end-to-end manner on the training data specifically prepared for the bi-directional loop closure task. Moreover, we also present an approach to estimate the relative pose between the query and the target image from the database, thus completing the chain of tasks required for comprehensive relocalisation. The 6-DoF pose regression in the proposed bi-directional case is significantly more challenging compared to traditional cases and is performed with an independent model.  We employ two public datasets namely FinnForest Dataset \cite{ali2020finnforest} and PennCOSYVIO dataset \cite{pfrommer2017penncosyvio} to conduct our tests for place recognition and pose regression.

\section{System Overview}
\begin{figure*}[!t]
\centering
\includegraphics[clip, trim=1.2cm 3.8cm 1.2cm 2.5cm, width=1.00\textwidth]{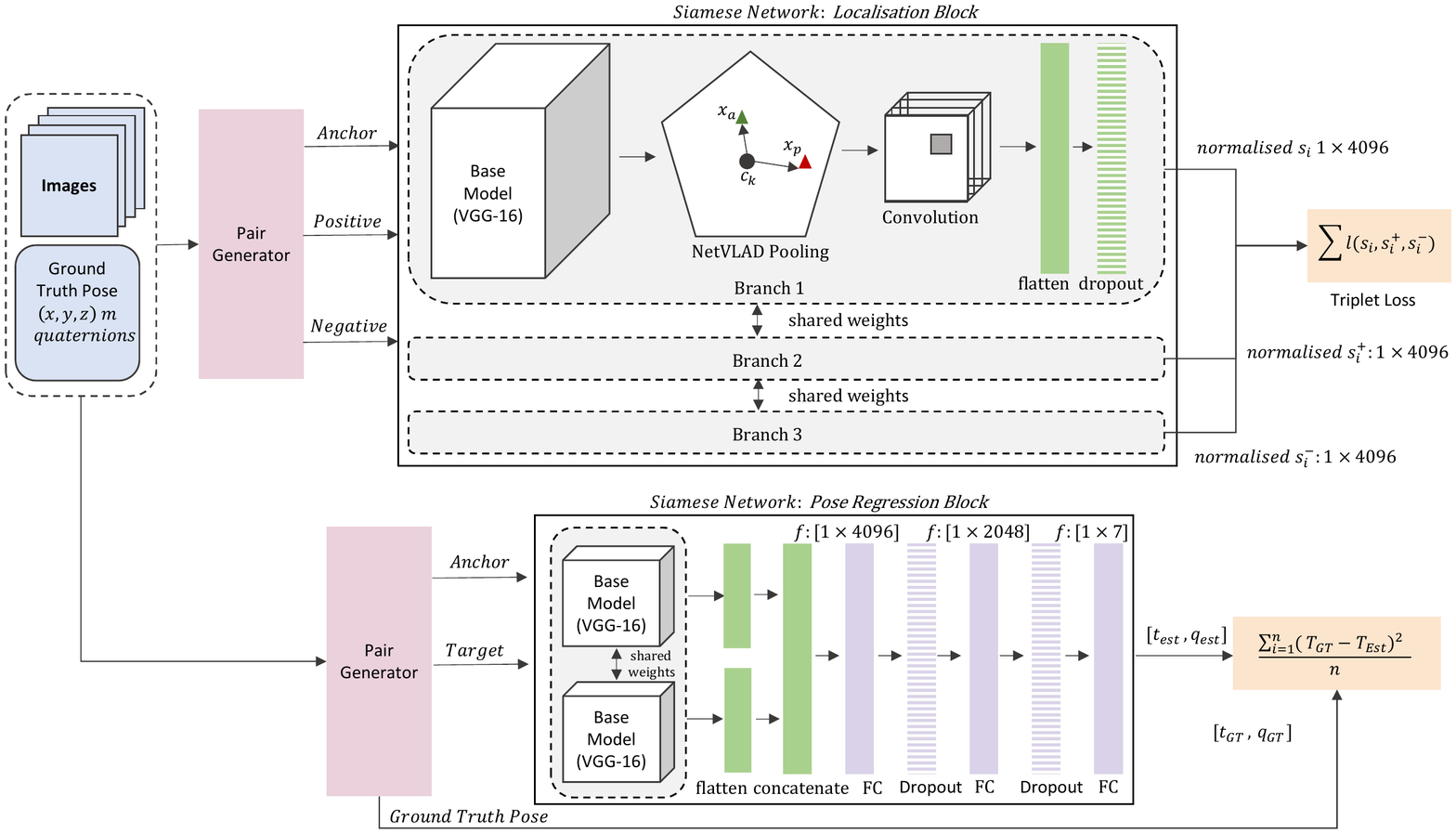}
\caption{ Illustration of the overall system pipeline. A siamese network constituting a VGG-16 base model topped with a VLAD pooling layer is used to learn similarity in the scenes using a triplet loss. Once the training process is completed,
we employ the branch as a descriptor to compute feature representations of database and query images for image retrieval towards localisation. The pose regression network (lower) is independently trained to directly regress the 6-DoF relative camera poses between the query and the retrieved match from the database.}
\label{pipeline}
\end{figure*}

In this section, we present the overall pipeline of the proposed approach for localisation. The approach constitutes of two modules: a siamese CNN network \cite{roy2019siamese} with triplet structure for maximizing similarity learning and a bi-input siamese model for 6 DoF relative camera pose regression. The overall pipeline is shown in Figure \ref{pipeline}. Initially, images are used from the database to train the siamese with a triplet structure for place recognition tasks by learning to identify maximum similarity. Each trained branch of the network is essentially a feature encoder and the extracted feature vectors can be employed to identify matches from the database of images that are nearest neighbors (NN) to the query image in the feature space. Afterward, samples with true positive matches are fed into the independently trained network for pose regression to estimate the relative pose between true matches. The processes are comprehensively explained in the following sections.

\subsection{Place Recognition}
\textbf{Feature Encoding} As stated earlier, we adopt a siamese approach for the task of place recognition whose illustration can be found in Figure \ref{pipeline}. The network constitutes of a base CNN model that takes three inputs namely query image sample \textit{\textbf{$I_q$}}, positive image sample \textit{\textbf{$I_p$}}, and negative image sample \textit{\textbf{$I_n$}} from the database \textit{\textbf{$I_D$}}. These input images are pre-processed based on the prescribed pre-processing technique adopted for the base model. Here, we take VGG-16 as an example which will be mainly used in our work; however, we also provide results with other base models for comparison later in the study. VGG-16 takes an input image of 224x224 pixels and propagates it through five sets of convolution and pooling layers, where the layers are connected through Rectified Linear Unit (ReLU) as an activation function. Each layer in the network learns a further abstraction of the input data with the highest-level abstraction residing towards the last layers. The structure is essential since features from the higher layers of the CNN hierarchy encode abstract semantic information about the scene, while features from intermediate and lower layers encode finer details from the image such as a change in appearance and structure \cite{sunderhauf2015performance}.

The outputs of the base model are normalized and fed into the neural network form of VLAD descriptor known as  NetVLAD pooling layer \cite{arandjelovic2016netvlad}. Essentially, VLAD encodes information about the statistics of local descriptors aggregated over an image in the form of feature distance from a cluster centre. For \textit{\textbf{$N$}} \textit{\textbf{$D$}}-dimensional local image descriptors \textit{\textbf{$\vec{x_i}$}} as input, and \textit{\textbf{$K$}} cluster centres (visual words) \textit{\textbf{$c_k$}} as VLAD parameter, the output VLAD image representation \textit{\textbf{$V$}} is \textit{\textbf{$K\times D$}}-dimensional. The L2-Normalized vector form of $V$ with elements $(j, k)$ is 

\begin{equation}
\label{ew_vlad}
V(j,k) = \sum_{i
=1}^{N} a_k(\vec{x_i}) (x_i(j) -c_k(j)).
\end{equation}
Here, $x_i(j)$ and $c_k(j)$ are the j-th dimensions of the i-th descriptor and k-th cluster centre, respectively. $a_k(\vec{x_i})$ indicates whether the descriptor $\vec{x_i}$  belongs to the k-th visual word, i.e. it is 1 if $\vec{x_i}$ belongs to the cluster $c_k$ and 0 otherwise.

To develop a layer reactive to training via backpropagation, it is required that the layer’s operation is differentiable with respect to all its parameters and the input. The original relation lacks this differentiation due to the binary nature of $a_k(\vec{x_i})$. To overcome this issue NetVLAD re-writes the original relation as:
\begin{equation}
\label{ew_netvlad}
V(j,k) = \sum_{i=1}^{N}        \frac{ e^{ w_k^T \vec{x_i} +b_k } } {\sum_{k{'}}   e^{ w_k{'}^T \vec{x_i} +b_k{'} }}        ( (x_i(j) -c_k(j)).
\end{equation}

where $w_k$ and $b_k$ are sets of trainable parameters for each cluster k which are learned in an end-to-end manner during training. Conceptually, the weight that the descriptor $\vec{x_i}$  is assigned to the cluster $c_k$ proportional to their proximity. Moreover, the relative proximity to other cluster centres also plays a part in the relation.

For our study, we found emperically that 64 clusters and 512-dimensional VGG16 backbone work effectively for the localisation task. The NetVLAD feature vector dimension becomes $512 \times 64 = 32, 768$. We further extract principle components through a convolution block and retrieve the encoded description as a normalized feature vector. 

\textbf{Loss Function} The similarity in an image is learned by employing a triplet loss over the output of each branch of the triplet siamese. For training, we gather training sample set $S$ such that
\begin{equation}
\label{ew_netvlad1}
S = \{(s_i, {s_i}^+,{s_i}^-) | (s_i, {s_i}^+ \in S^+) ;(s_i, {s_i}^- \in S^- ), i=1,...,M\}.
\end{equation}

Here, $S^+$ refers to the set of relevant image pairs, $S^-$ refers to negative  image pairs, and M indicates the span of the entire triplet set. The triplet loss is then  given as

\begin{multline*}
\label{ew_netvlad}
\ell(s_i, {s_i}^+,{s_i}^-) = \\
max(0,m + \left \| f(s_i)-f({s_i}^+) \right \| - \left \| f(s_i) - f({s_i}^-) \right \|)
\end{multline*}

Here, margin $m$ is a scaler that defines an offset between positive and negative pairs, and $f(.)$ is an embedding of an image sample. The global loss over all triplet samples is given as 
\begin{equation}
\label{ew_netvladsum}
L= \sum\limits^{}_{(s_i, {s_i}^+,{s_i}^-) \in \textit{\textbf{S}}}   \ell(s_i, {s_i}^+,{s_i}^-)
\end{equation}

\textbf{Retrieving the nearest neighbours}
To retrieve a potential match for a query image from the database of images, both images must have a suitable representation before comparison. In the proposed case,  we use one branch of the fine-tuned network as a feature extractor to encode the query and database images. This enables us to have the representation in the same embedding space (i.e. 4096-dimensional feature vectors, see
Figure \ref{pipeline}). In the experimentation section, we will use other methods of feature extraction to encode our images for the sake of comparison. Finally, the top N-ranked database images, $d = (d_n|d_n \in D, n = 1...N)$ are selected as the nearest neighbors to the query image based on the squared Euclidean distance in the embedding space. It is important to note that a query image might have one, many, or no match in the database as it depends on the number of keyframes generated during earlier exploration of the environment.

\textbf{Neighbour Confidence Sharing}
Place recognition is a critical task for loop closure in visual SLAM. The problem can become significantly more challenging when the environment contains repetitive textures even for distinct locations such as in forests and large open areas. It is often the case that wrong matches are generated due to similar semantics of different scenes. To overcome this problem, we propose a confidence sharing scheme where the confidence of the previously localized points is propagated to their neighbors in a causal manner. In our case, we consider three neighbors for sharing confidence. We incorporate the traveled distance between the neighbors in order to ascertain the sanity of a potential match for a query point. A new query point has a valid match if (1) it has a considerable match score (in embedding space from the model) with an image from the database and (2) it has nearby localized neighbors that agree in distance traveled with the estimates from odometry. If a new neighbor is found far away from a nearby localized neighbor and the odometry estimates run in favor of the previously localized neighbor, then the new match is discarded as a possible wrong match.

\subsection{Pose Regression}
\label{poseReg_sec}
% \textcolor{red}{STRESSON NEED OF POSE REGRESSOR USING END-END-Learning}
The pose estimator block composes of a VGG-based siamese architecture that takes two monocular images as raw input and predicts a 6-DoF relative transformation between the poses for those specific inputs. The siamese regression block is shown in Figure \ref{pipeline}. The shared weights are initialized with a network pre-trained for large-scale place classiﬁcation task \cite{zhou2017places} using the Places 365 dataset, and later ﬁne-tuned for the relative pose estimation task as described below. The output of each branch is vectorized and combined into a single encoded description. The relative pose is regressed by passing the feature vector through three fully connected (FC) layers activated through Leaky ReLu functions and intermediate dropout layers for regularization, as shown in Figure \ref{pipeline}. The final FC layer gives out the predicted relative pose. 
Different studies adopt different representations for the angles. In study \cite{jau2020deep}, the authors use deeply learned key points to estimate the Fundamental matrix between the two images using two CNN modules. In a similar study titled UnDeepVO \cite{li2018undeepvo}, the authors opt for direct image alignment in conjunction with the camera intrinsic to estimate the relative pose. The relative pose is obtained as a decoupled combination of a translation vector and rotation vector in Euler angles.

Although Euler angles carry briefer representation compared to the fundamental matrix, however, it suffers from discontinuities in the form of gimbal lock. On the other hand, rotation parameterization such as rotation matrices that lie on a manifold, their distance computation requires ﬁnding a Euclidean embedding. In our work, we represent the angles using quaternions similar to the work \cite{laskar2017camera}.  It is important to note that quaternions lie on a unit sphere, however, during optimization/training the difference becomes so small that the distinction between spherical distance and Euclidean distance becomes insigniﬁcant. Therefore, to avoid obstructing the optimization with unnecessary additional constraints, we avoid the use of spherical constraint. Hence, the distance between two quaternions can be measured by the Euclidean l2 norm $\left \|q_{GT}- q\right \|$. The authors of popular study PoseNet \cite{kendall2015posenet} and its derivative study \cite{laskar2017camera} argue to use a decoupled approach with a weighted parameterization of the angle, with a scale factor $\beta$, to balance the loss function

\begin{equation}
\label{reg_l_1}
L = \left \| \Delta t_{GT} - \Delta t \right \|^2+ \beta \left \| \Delta q_{GT} -\Delta q \right \|^2.
\end{equation}

During experimentation, we observed that the approach seemed cumbersome as the vale of $\beta$ has to be manually adjusted for each dataset. The authors \cite{kendall2015posenet} remark that the value of $\beta$ can lie anywhere in the range of 120 to 2000 depending on the structure and semantics of the scene \cite{kendall2015posenet}. To avoid this issue we propose to discard the scale factor $\beta$ and independently scale down the entire translation vector and quaternions to the same range during the preprocessing step of dataset preparation. Since we wish to use an adaptation of ReLu activation for the FC layer, it is advisable to rescale both the quaternions and the translation vectors between $[0,1]$. This makes the relation invariant of any scale factor for the training phase. The scale factors can be extracted from the range of the data in the dataset and applied using relation 

\begin{equation}
\label{eq_scale}
d_{scaled}= \frac{(sc_{max} - sc_{min}) * (d - d_{min})} {d_{max} - d_{min}}.
\end{equation}
where $d$ denotes the data array and $sc$ indicates the scaler values of the desired range for scaling. The model is trained to predict an arbitrarily scaled version of the pose where the scale is restored in a post-processing step after the prediction. The MSE is then given as 

\begin{equation}
\label{reg_l_2}
L_{MSE} = \frac{1}{n} \sum_{i=1}^{n}  (T_{scaled \; GT} - T_{Est.})^2.
\end{equation}

Here, $T_{scaled \; GT}$ is the pose constituting scaled $[t_x,  t_y,  t_z]$ and scaled $[q_w, q_x, q_y, q_z]$. $T_{Est.}$ is a similar vector to $T_{scaled GT}$ which is predicted by the model. At test time, a pair of images are fed into the regression model, consisting of two branches, which directly estimates the relative camera pose vector. Finally, the estimated quaternion and translation vectors are scaled up to retrieve real-world values.

\subsection{Data Selection and Dataset Preparation}
\label{data_Sel_SEC}
Deep learning approaches require a large amount of data in order to train a model that can generalize well for all cases. Producing new data every time is often expensive and can drive the researcher away from the problem at hand. For this reason, many public datasets have been introduced by researchers in their respective fields. Visual SLAM is a widely researched problem and many datasets exist for testing purposes \cite{engel2016photometrically, carlevaris2016university, zhu2018multivehicle, jung2016multi, maddern20171}. However, most of the datasets do not provide bi-directional motion since they are tailored for handling the problem from a uni-directional perspective. For our work, we found that FinnForest dataset \cite{ali2020finnforest} and parts of PennCOSYVIO dataset \cite{pfrommer2017penncosyvio} can be used for training and testing purposes.

As indicated by the name, the FinnForest dataset provides data, for visual odometry and SLAM, in a forest landscape. The dataset provides recordings from four RGB cameras that are synchronized with an Inertial Measurement Unit (IMU), and a Global navigation satellite system (GNSS). The dataset contains sequences for odometry that are well suited for this study. Each route is traveled from both directions within the same sequence thus providing all the relevant data for bi-directional loop closure. The dataset is challenging for the problem since it contains repetitive texture, unlike an urban landscape that provides more distinct landmarks over its trajectory. We will only utilize the data recorded in the summer conditions in our study since it offers slightly more landmarks than the winter condition for the localisation block. 

The second dataset that we use for training and testing offers indoor data. This was specifically chosen to check the performance of our approach in both indoor and outdoor environments. The dataset records similar sequences with multiple configurations and types of sensors. We found four sequences, recorded with GoPro Hero 4 Black, that are reasonably well suited for the task of bi-directional loop closure. The sequences include slow and fast motions that can represent vehicular motion using a forward-facing camera. Moreover, the data includes ground truth poses that can be used for automatic extraction of training and testing samples in our approach. Other sequences in this dataset are wall-facing and more suited for Structure from Motion applications.
 
 Both of the datasets are passed through a data preparation phase in order to generate sequences that can be used for training our localisation and pose regression models. We generate sub-datasets out of the original datasets and use them for training. Since the localisation and the pose regression are to be performed on the same scenes we can use the data generated for localisation in the pose regression block. For simplification, we split and consider two cases of the bi-directional localisation problem. Assume that a route is traversed in a straight line from both directions then we have images acquired with a camera at somewhat regular intervals from both directions for roughly the same location, given that the camera frame rate is high enough. 
 
\begin{figure}[h]
\centering
\includegraphics[clip, trim=0cm 13cm 7.5cm 2.2cm, scale=0.48]{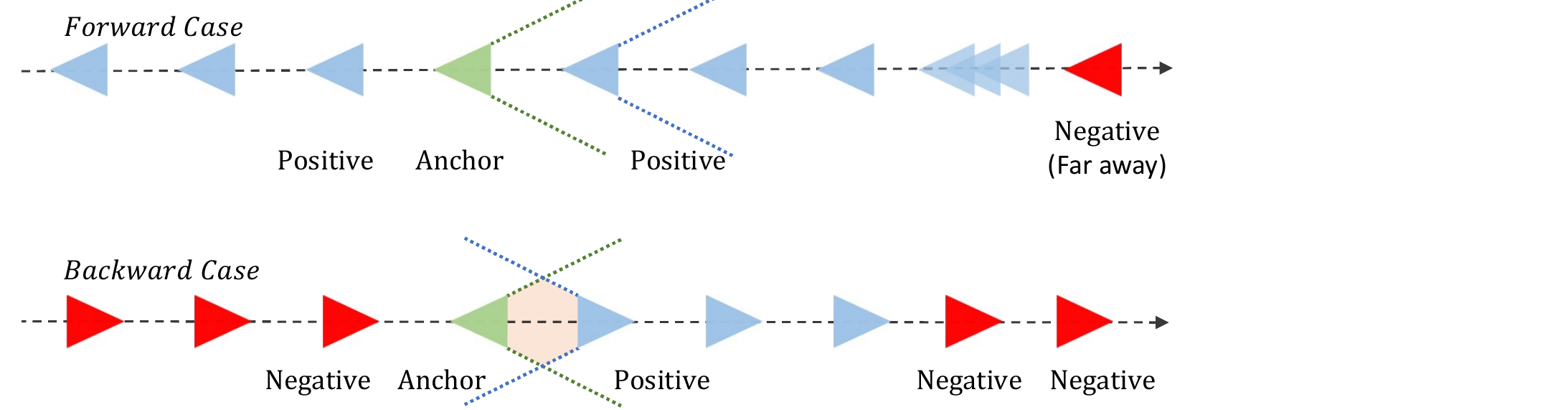}
\caption{ Illustration of training data selection based on distance and direction. The cones represent camera body placed at various locations. The shaded area in the backward case represents potentially overlapping regions in the perspective view}
\label{Datascheme}
\end{figure}

 Considering the forward motion case, in Figure \ref{Datascheme}, an anchor sample is acquired at query location (green). For these anchor samples, we can obtain positive sample pairs from nearby locations that share the perspective view. Moreover, it is fairly safe to assume that an image acquired further away or from far back will provide a significantly different view and can be selected as negative samples (red) for the localisation model training. For all the sample pairs we compute the relative ground-truth poses in the form of translation vector and quaternion angles which are later used in the pose regression block. We generate 6 sample triplets for each query location, however, any number of triplets can be generated according to the needs of the task. The straight-line route shown in Figure \ref{Datascheme} is a simplified case, however, we should expect irregular movements and changes in angles in real cases. The FinnForest dataset attempts to mimic the conditions that a heavy vehicle might face in a real forest during its operation. we expect close-by samples from a query point that might not share the same visual information due to sharp turns, camera jitter, motion blur, sudden overexposure, or sun flare in the camera view. These samples can deteriorate the learning performance of the model as it expects them to be positive samples however they no longer share the semantic and/or geometric information with the query sample.

 To overcome this, we leverage structure from motion to autonomously generate training triplets by validating the previously created triplet samples. We generate 3D world points from a query stereo pair and track the corresponding key points among all the positive samples for that specific query sample. Similarly, the 3D points are also propagated between the camera frames using the 6 DoF ground-truth poses. The transformed world points are then projected into corresponding image space. Any sample with  cumulative reprojection error (for the tracked keypoints) higher than a threshold is discarded. The reprojection error is computed using 
 
 \begin{equation}\label{dataSeleRerpoj}
\begin{split}
E_{px} =  || P_{s_i^+} - \Pi (K, [q_{(s_i,s_i^+)}\:,\:_{s_i}t^{s_i^+}]_{HT} , W_{s_i}) ||_2^2   .
\end{split}
\end{equation} 
 
 This validation step gets rid of the sample pairs with high angular changes in perspective (such as in the case when the vehicle is turning). Here, $\Pi$  is the perspective projection function that projects the 3D points $W=(X,Y,Z,1)^T$ from world frame space to image space using the camera intrinsic $K$. The superscript $T$ indicates the transpose of a vector.  The perspective projection yields $\tilde{x}=(\tilde{u},\tilde{v},1)^T$ in the image space of the camera at the pose of interest. The reprojected points $\tilde{x}$  are compared directly against the observed/tracked 2D points ($P$) in the corresponding sample image. The symbol $[\:]_{HT}$ indicates the conversion from quaternion $q$ and translation vector $t$ to the homogeneous transformation matrix. We use quaternion angle representation for the sake of coherence.

For the backwards case, We use the same query image samples initially selected and filtered for the forward case and attempt to find pairs for it in the backward motion part of the sequence. In contrast to the forward case, the backward motion case can have positive samples only ahead of the query location (see Figure \ref{Datascheme}). All the potential samples are expected to be oriented in the direction opposite to the camera orientation at the query point. Moreover, the assumption is that camera poses that are slightly ahead of the query point would share potentially more of the same scene even if from the opposite perspective. The similarity in the scene in this small range is what we want our model to learn and discriminate. A camera pose that is too far ahead or at the back of the query pose would have little and no match with the query perspective, respectively. As before the positive samples are indicated in blue while negative samples are shown in red.  The reprojection-based verification is not possible for the backward case since the traditional feature detector cannot detect and track features with such high perspective changes. As a result, there are no reference key point image positions for the reprojected 3D points.
 
 \begin{figure}[h]
\centering
\includegraphics[scale=0.65]{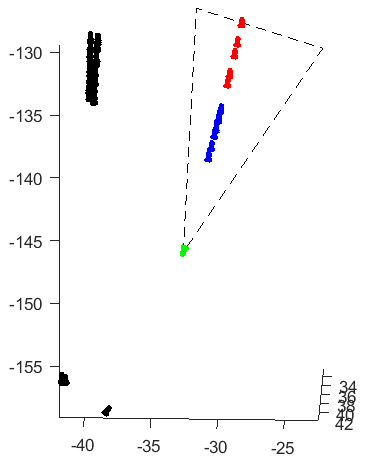}
\caption{Visualization of the automatic sample selection scheme for the case of backward motion. Valid training samples within the field of view of the anchor/query pose (green) are shown in blue while the negative samples are shown in red. The poses shown in black indicate rejected samples in the vicinity of the query.}
\label{neg_sampSel}
\end{figure}

A visualization of the automatic sample selection, based on the camera poses, from a sequence of FinnForest dataset is shown in Figure \ref{neg_sampSel}. Samples with in the field overview of the query pose are considered as potential candidates for triplet grouping. Similarly, we discard very close samples since target samples that are too close will share very little view with the query perspective. Here, the camera poses are shown that pass the constraints set on the field of view, distance from the query pose, and the tolerance of orientation difference from the query pose orientation. The candidates for positive samples are shown in blue while the negative samples are shown in red. Additionally, some camera poses that do not pass the constraints are visualized in black for the sake of understanding. 

\begin{figure}[!t]
\centering
\includegraphics[scale=0.65]{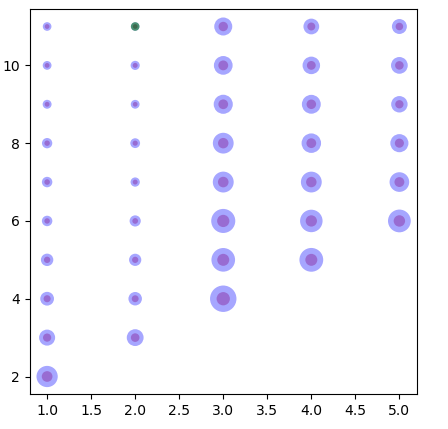}
\caption{Upto scale visualization of the training loss (purple) and generalization gap (indigo) when the training data is changed for the backward case. The horizontal axis indicates the minimum valid distance i.e, the distance from query to the nearest sample  while the vertical axis indicates the maximum valid distance i.e, the distance from query to the furthest valid sample. We indicate the best result with a green overlay.}
\label{RangeLossGap}
\end{figure}

It is difficult to conclude from merely visually observing the data as to what should be the minimum and maximum distance between the query and the positive samples. To understand the relationship we follow an empirical approach and train the model with different data distributions. In Figure \ref{RangeLossGap}, we explain the effect of data distribution. The figure shows an Upto scale visualization of the training loss (purple) and generalization gap (indigo) for the backward case. The horizontal axis indicates the minimum valid distance i.e, the distance from the query to the nearest sample while the vertical indicates the maximum valid distance i.e, the distance from the query to the furthest valid sample. We observed that the training loss and the generalization gap were minimum for the training data when the nearest positive sample was limited to a distance of 2 meters and the farthest sample was kept to be at 11 meters from the query pose. This indicates that the maximum overlap is found within this range for the given data and that anything out of this range is a potential outlier to what the model attempts to learn. This is obviously specific to the datasets in this study and might vary slightly depending upon the camera and optics used for recordings.

\section{Results and Discussion}
% \textcolor{red}{Localisation and regression together did not work}
In this section, we provide our experimental results and quantitatively demonstrate the effectiveness of the proposed system on the FinnForest and PennCOSYVIO datasets. To ascertain the generalization capability of our pipeline, on data previously unseen during training, we hold out one of the scenes in the FinnForest dataset (S5)  and PennCOSYVIO dataset (C2-bs) for evaluation and train our model on the remaining scenes. We will discuss the results in the forthcoming subsections and compare our results with other well-established methods.
The network models were implemented with the Tensor Flow framework using Keras API. We employ the Adam optimizer to train the network with an early stop. The network setup preferred a small learning rate that started from .0000001 and decreased by one-tenth every epoch.

\subsection{Place Recognition}
For testing place recognition capability, we compare our proposed method with three other approaches to gauge the relative quality of the results. Among these methods, VGG-FC and VGG FC-norm are variants of deep learning approaches where we deploy two fully connected layers with dropouts applied after the VGG-16 network. VGG FC-norm has an additional normalization layer before the feature encoded vectors are extracted from the network.  The third approach which we term here as ORB-VLAD uses ORB feature detector and descriptor to encode keypoints from the images and uses VLAD to further re-encode and reduce the dimensionality of the feature vectors. The use of VLAD helps us to have a more direct comparison with our proposed approach since we employ a variant of VLAD known as netVLAD.

\begin{figure}[h]
\centering
\subfloat[]{\includegraphics[clip, trim=2.3cm 7cm 2.7cm 7.8cm, scale=0.48]{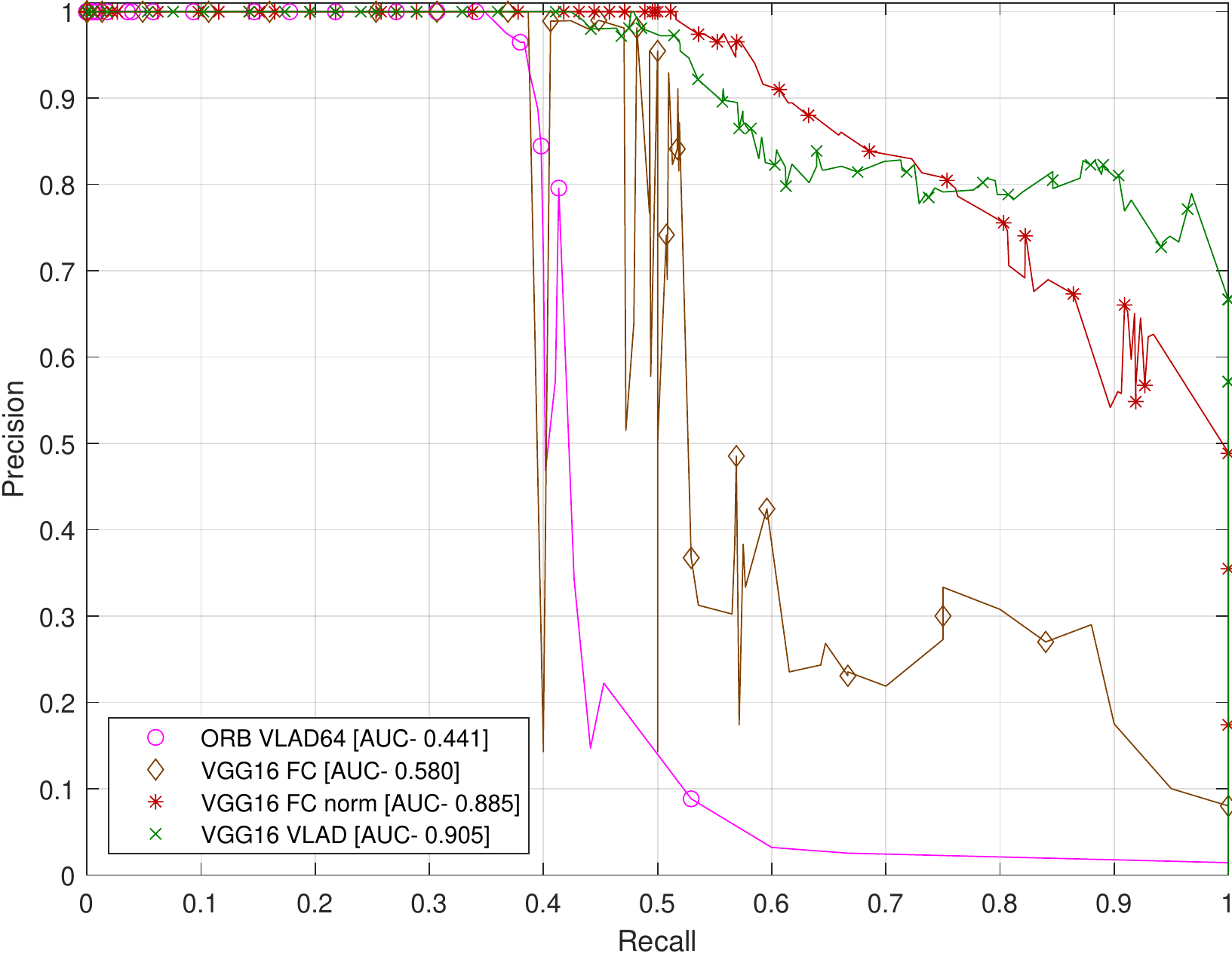}%
\label{pr_localisation_finn}}
\hfil
\subfloat[]{\includegraphics[clip, trim=2.3cm 7cm 2.7cm 7.8cm, scale=0.48]{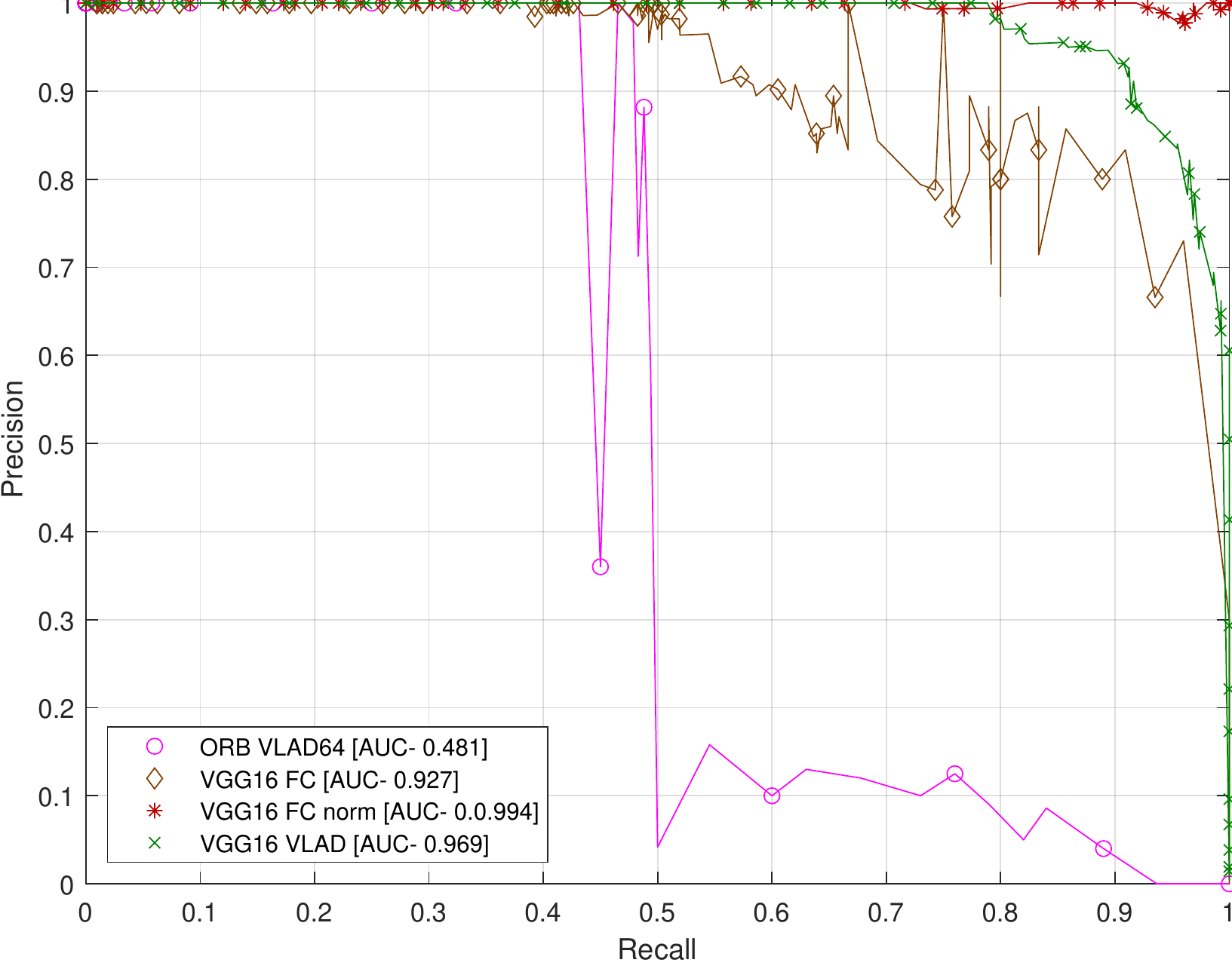}%
\label{pr_localisation_penn}}
\caption{Precision-recall curves for bi-directional loop closures in the (a) FinnForest dataset and (b) PennCOSYVIO dataset.}
\label{PR_ Localisation}
\end{figure}

For both datasets, the localisation performance is expressed in the form of a Precision-Recall (PR) curve. The results are shown in Figure \ref{PR_ Localisation}.
It can be observed from the results that the proposed approach VGG16-VLAD and VGG16-FCnorm outperform VGG16-FC and ORB-VLAD. The difference between the area under the curve (AUC) for VGG16-VLAD and VGG16-FCnorm in case of both datasets is almost the same. Nonetheless, we remark that VGG16-VLAD is more suited for the task at hand. Our proposition is based on the observation that VGG16-VLAD performs better on the FinnForest dataset which is considerably more challenging compared to the PennCOSYVIO dataset. FinnForest dataset is recorded over a significantly larger spatial area which has repetitive textures and fewer discriminative landmarks. On the other hand, the PennCOSYVIO dataset offers the same indoor scene in all sequences where the route is the same and motion speed is slightly varied. This means we can expect a high correlation in the training and testing data in the case of PennCOSYVIO dataset. In contrast, the route and scene are varied in the FinnForest dataset which will result in a lower correlation between training and testing data, and higher data center distribution (in space that houses encoded data clusters). Hence, we can infer that VGG16-VLAD has better generalization capability compared to the other methods. It is important to mention that we also tested the SURF and SIFT features in combination with VLAD. However, the results were not included as they were poor and inhibited the readability of the PR curve. These classical feature descriptors work well for the uni-directional cases where the perspective does not change a lot, however, they fail to perform well in the bi-directional cases.

\begin{figure}[t!]
\centering
\subfloat[]{\includegraphics[clip, trim=3cm 1cm 3cm 2cm, scale=0.28]{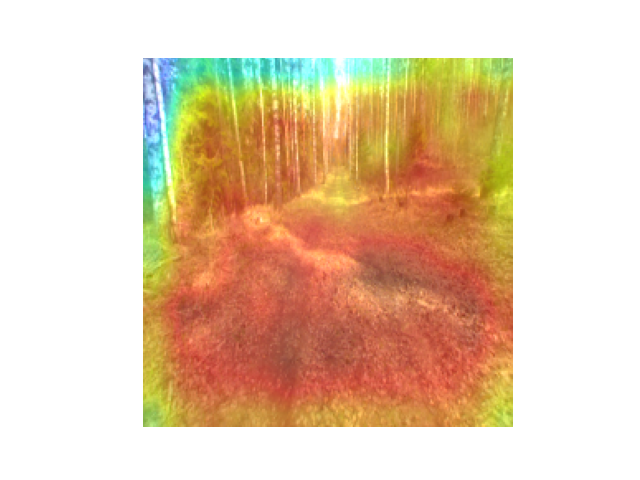}}
\subfloat[]{\includegraphics[clip, trim=3cm 1cm 3cm 2cm, scale=0.28]{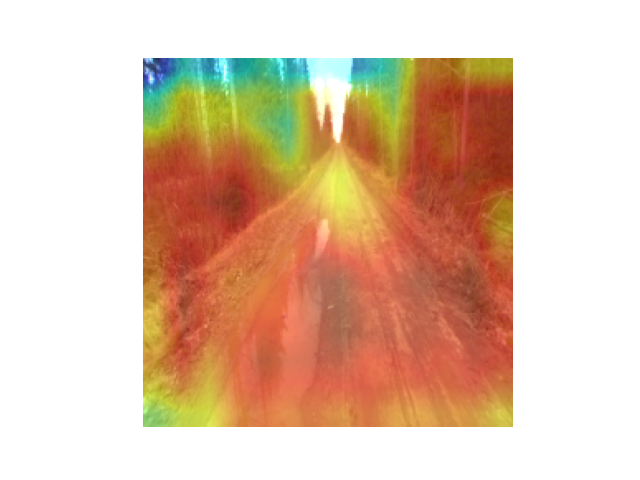}}
\subfloat[]{\includegraphics[clip, trim=3cm 1cm 3cm 2cm, scale=0.28]{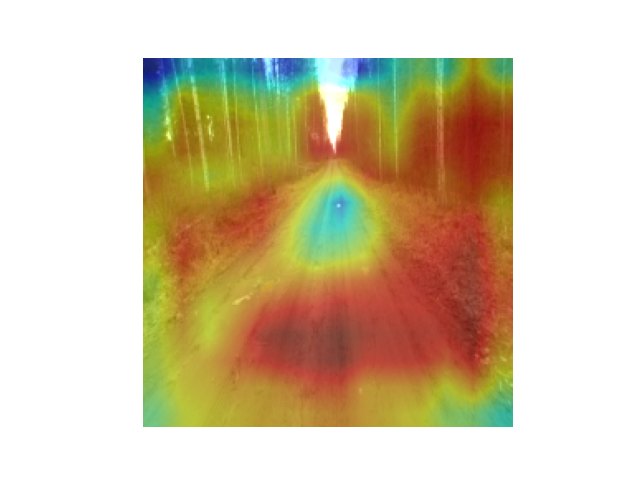}}

\subfloat[]{\includegraphics[clip, trim=3cm 1cm 3cm 2cm, scale=0.28]{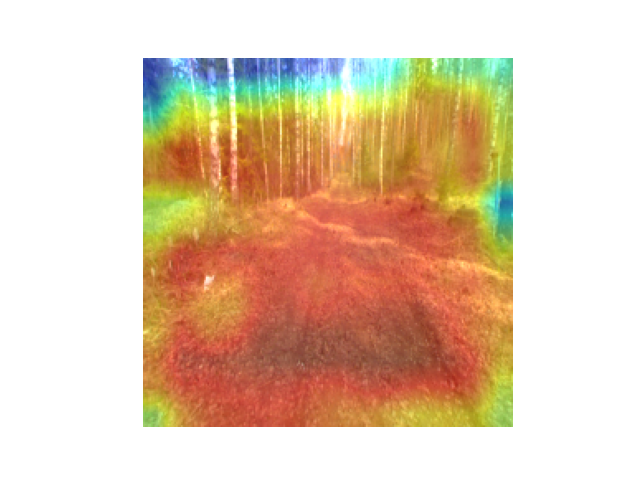}}
\subfloat[]{\includegraphics[clip, trim=3cm 1cm 3cm 2cm, scale=0.28]{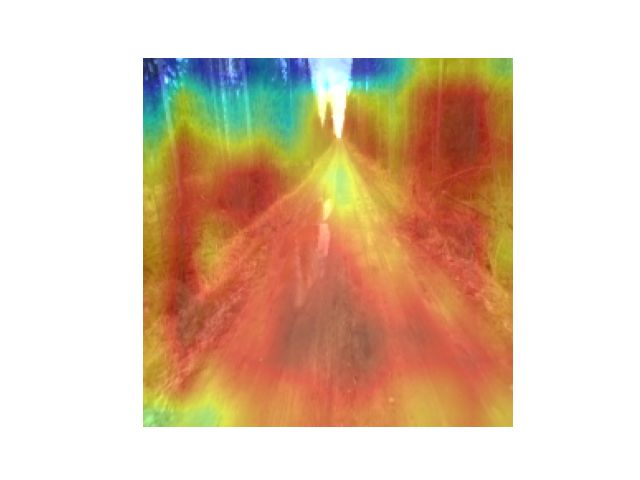}}
\subfloat[]{\includegraphics[clip, trim=3cm 1cm 3cm 2cm, scale=0.28]{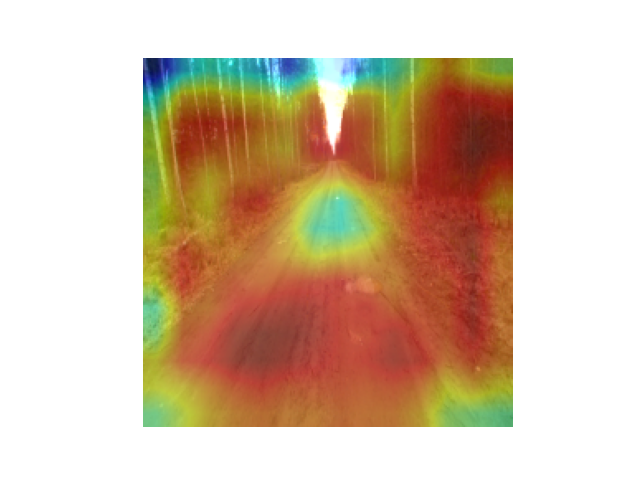}}

% \subfloat[]{\includegraphics[clip, trim=3cm 1cm 3cm 2cm, scale=0.28]{Figures/seq1_10295_n.png}}
% \subfloat[]{\includegraphics[clip, trim=3cm 1cm 3cm 2cm, scale=0.28]{Figures/seq3_1213_n.png}}
% \subfloat[]{\includegraphics[clip, trim=3cm 1cm 3cm 2cm, scale=0.28]{Figures/seq5_6389_n.png}}

\caption{Activation maps over sample image triplet used for testing from FinnForest Dataset for forward/uni-directional case, where maps in (a-c) are for query images, (d-f) are for corresponding (column wise) positive samples.}
% , and (g-i) are for the negative samples}
\label{act_finn_for}

\centering
\subfloat[]{\includegraphics[clip, trim=3cm 1cm 3cm 2cm, scale=0.28]{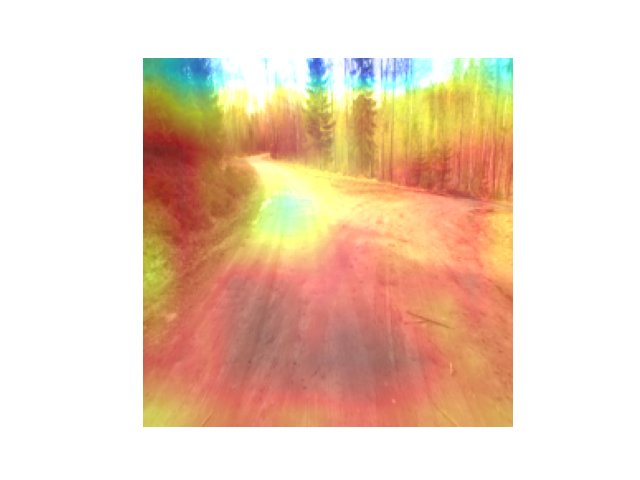}}
\subfloat[]{\includegraphics[clip, trim=3cm 1cm 3cm 2cm, scale=0.28]{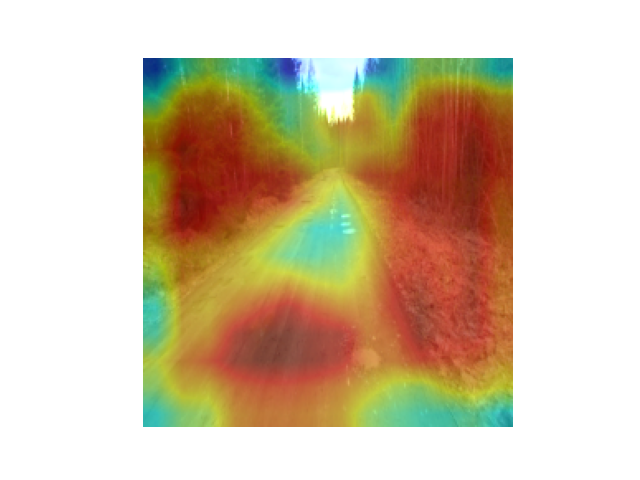}}
\subfloat[]{\includegraphics[clip, trim=3cm 1cm 3cm 2cm, scale=0.28]{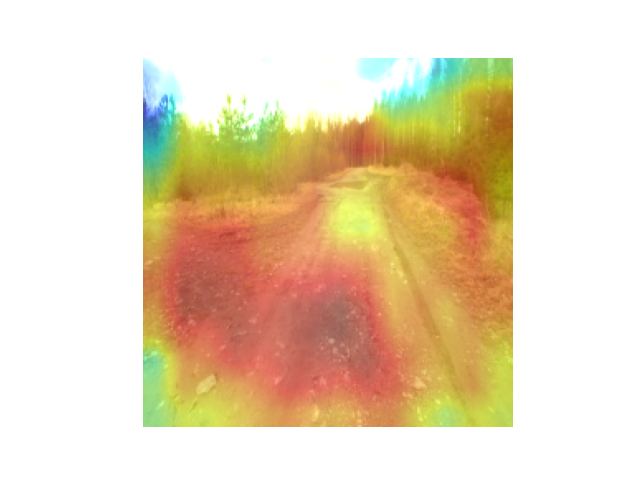}}

\subfloat[]{\includegraphics[clip, trim=3cm 1cm 3cm 2cm, scale=0.28]{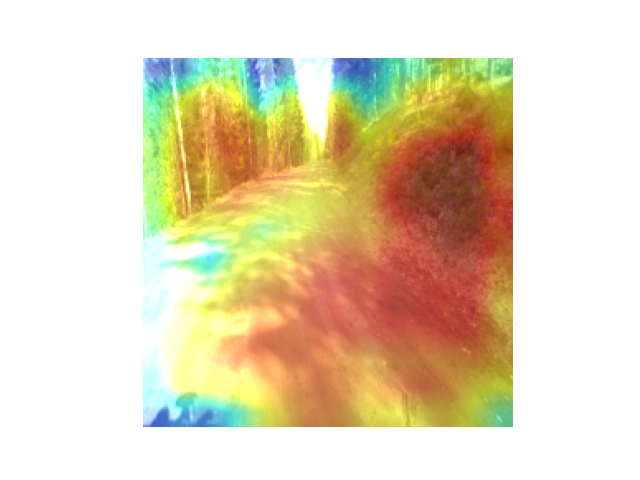}}
\subfloat[]{\includegraphics[clip, trim=3cm 1cm 3cm 2cm, scale=0.28]{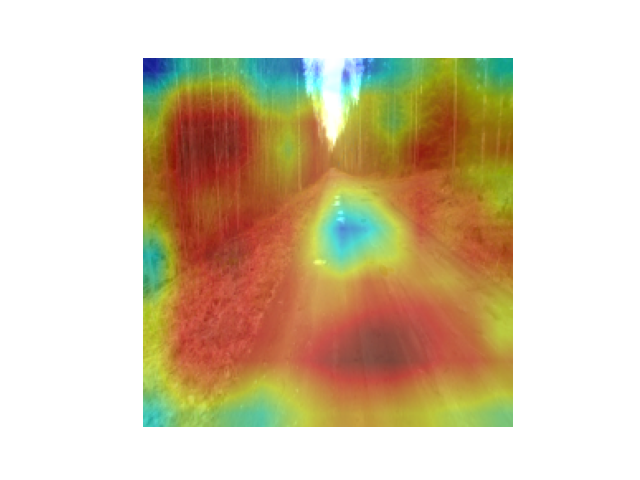}}
\subfloat[]{\includegraphics[clip, trim=3cm 1cm 3cm 2cm, scale=0.28]{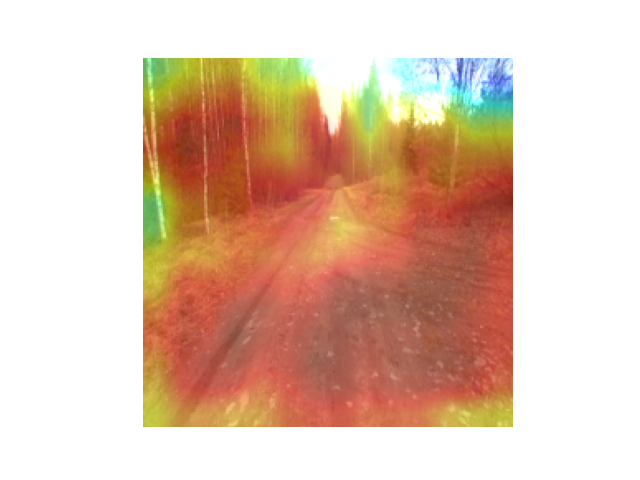}}

% \subfloat[]{\includegraphics[clip, trim=3cm 1cm 3cm 2cm, scale=0.28]{Figures/seq1_8878_n.png}}
% \subfloat[]{\includegraphics[clip, trim=3cm 1cm 3cm 2cm, scale=0.28]{Figures/seq5_4046_n.png}}
% \subfloat[]{\includegraphics[clip, trim=3cm 1cm 3cm 2cm, scale=0.28]{Figures/seq5_26815_n.png}}

\caption{Activation maps over sample image triplet used for testing from FinnForest Dataset for bi-directional motion case, where maps in (a-c) are for query images, (d-f) are for corresponding (column wise) positive samples}
% , and (g-i) are for the negative samples}
\label{act_finn_rev}
\end{figure}

\begin{figure}[t!]
\centering
\subfloat[]{\includegraphics[clip, trim=3cm 1cm 3cm 2cm, scale=0.28]{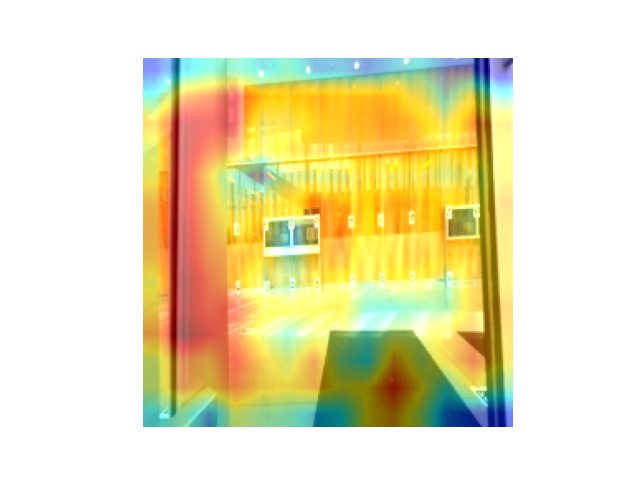}}
\subfloat[]{\includegraphics[clip, trim=3cm 1cm 3cm 2cm, scale=0.28]{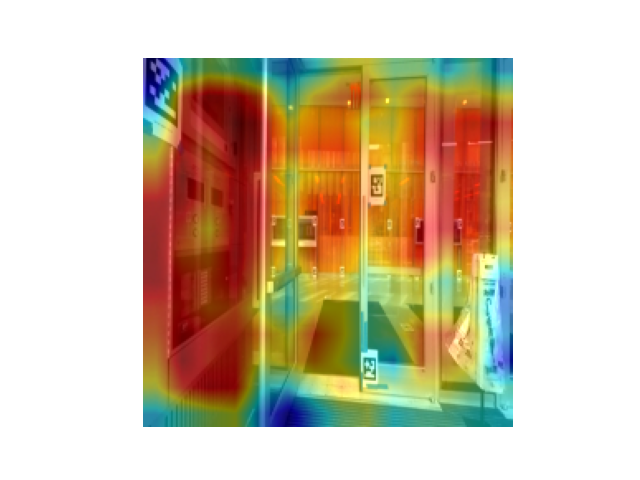}}
\subfloat[]{\includegraphics[clip, trim=3cm 1cm 3cm 2cm, scale=0.28]{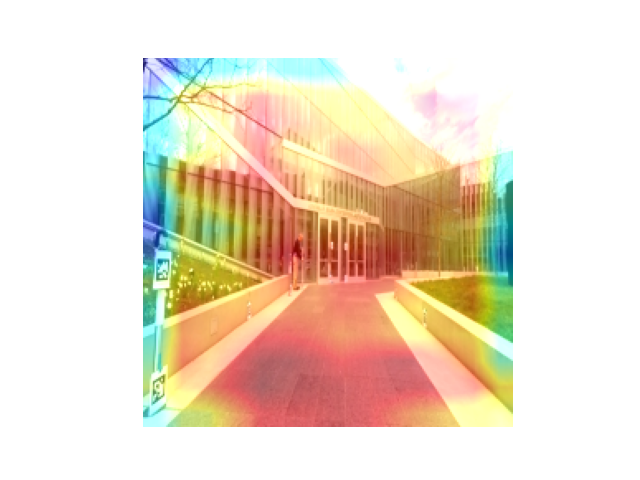}}

\subfloat[]{\includegraphics[clip, trim=3cm 1cm 3cm 2cm, scale=0.28]{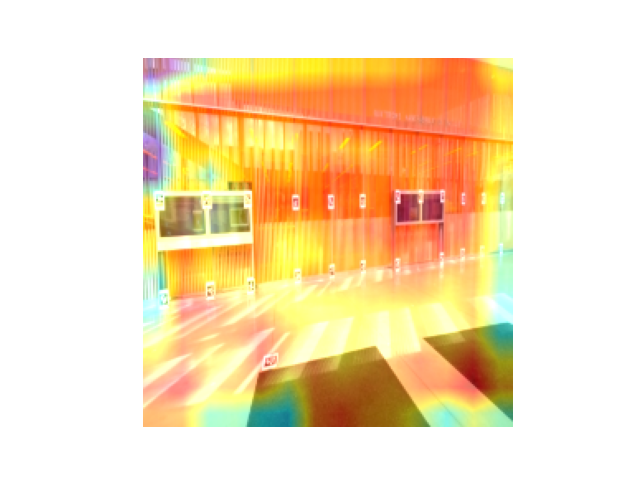}}
\subfloat[]{\includegraphics[clip, trim=3cm 1cm 3cm 2cm, scale=0.28]{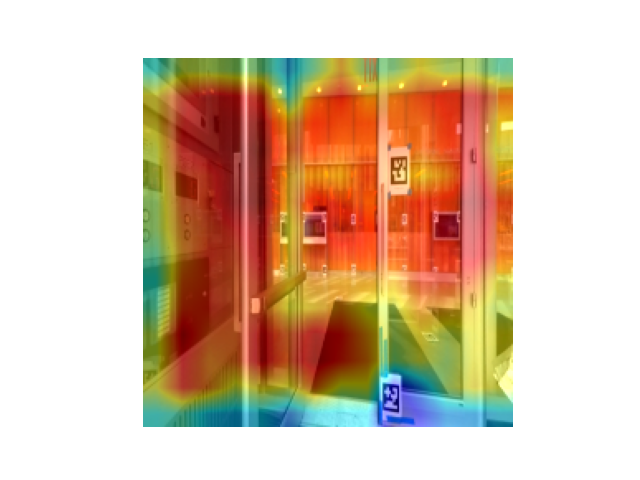}}
\subfloat[]{\includegraphics[clip, trim=3cm 1cm 3cm 2cm, scale=0.28]{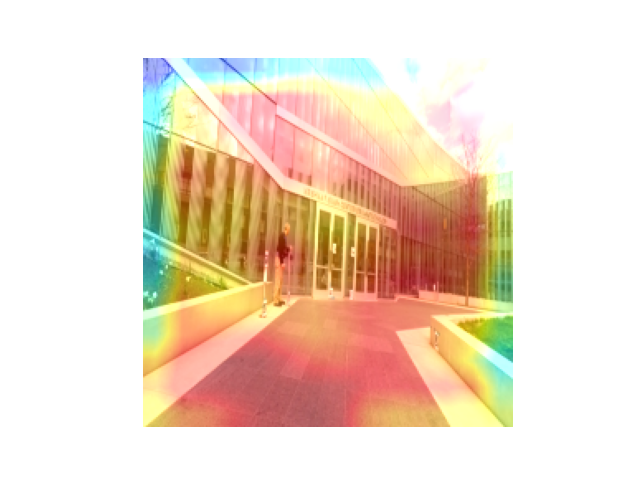}}

% \subfloat[]{\includegraphics[clip, trim=3cm 1cm 3cm 2cm, scale=0.28]{Figures/af_716_n.png}}
% \subfloat[]{\includegraphics[clip, trim=3cm 1cm 3cm 2cm, scale=0.28]{Figures/af_651_n.png}}
% \subfloat[]{\includegraphics[clip, trim=3cm 1cm 3cm 2cm, scale=0.28]{Figures/af_430_n.png}}

\caption{Activation maps over sample image triplet used for testing from PennCOSYVIO Dataset for forward/uni-directional motion case, where maps in (a-c) are for query images, (d-f) are for corresponding (column wise) positive samples.}
% , and (g-i) are for the negative samples}
\label{act_penn_for}

\centering
\subfloat[]{\includegraphics[clip, trim=3cm 1cm 3cm 2cm, scale=0.28]{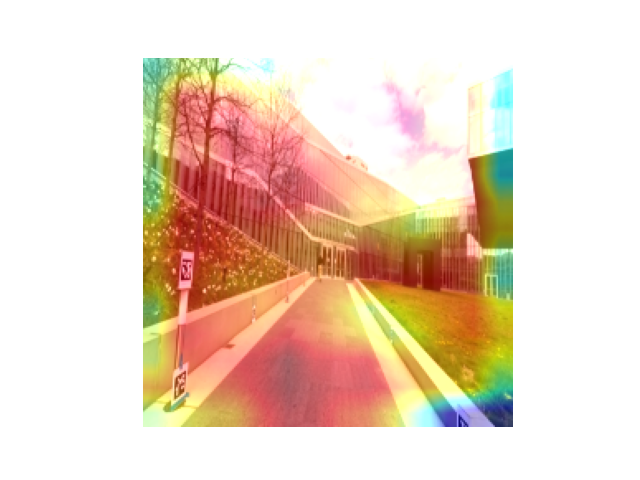}}
\subfloat[]{\includegraphics[clip, trim=3cm 1cm 3cm 2cm, scale=0.28]{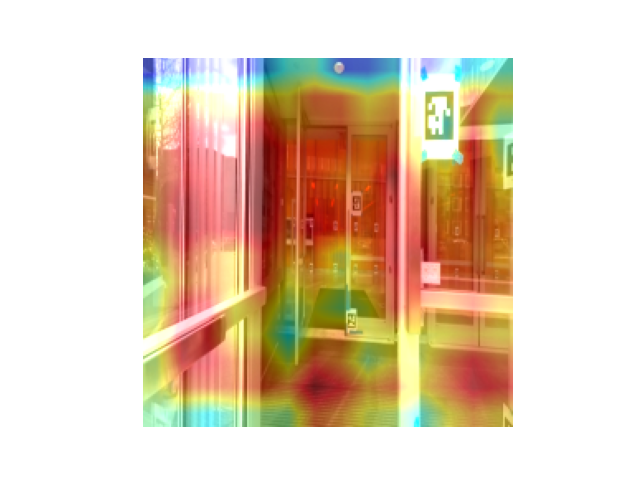}}
\subfloat[]{\includegraphics[clip, trim=3cm 1cm 3cm 2cm, scale=0.28]{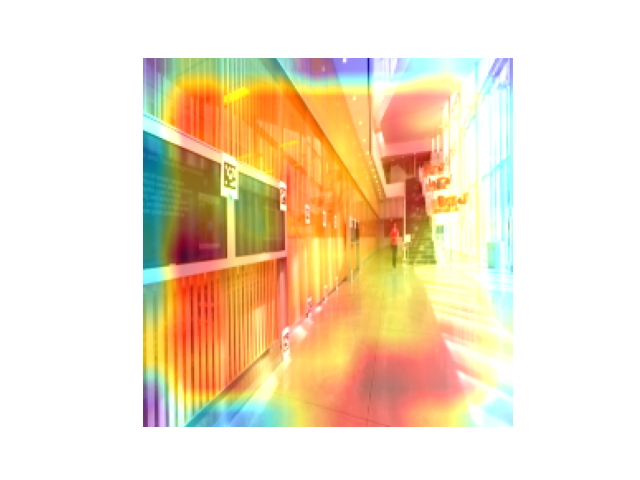}}

\subfloat[]{\includegraphics[clip, trim=3cm 1cm 3cm 2cm, scale=0.28]{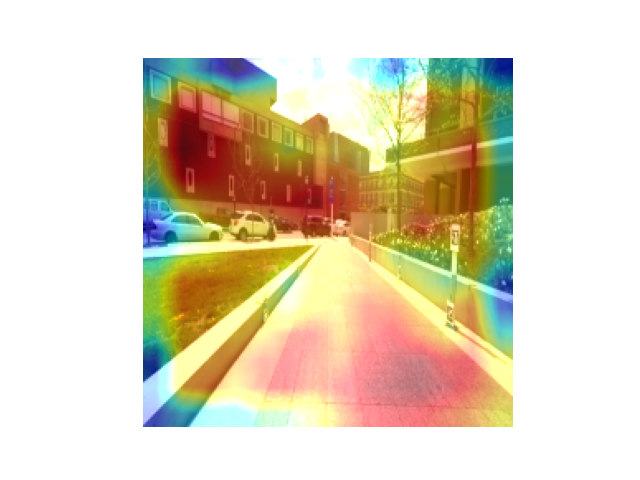}}
\subfloat[]{\includegraphics[clip, trim=3cm 1cm 3cm 2cm, scale=0.28]{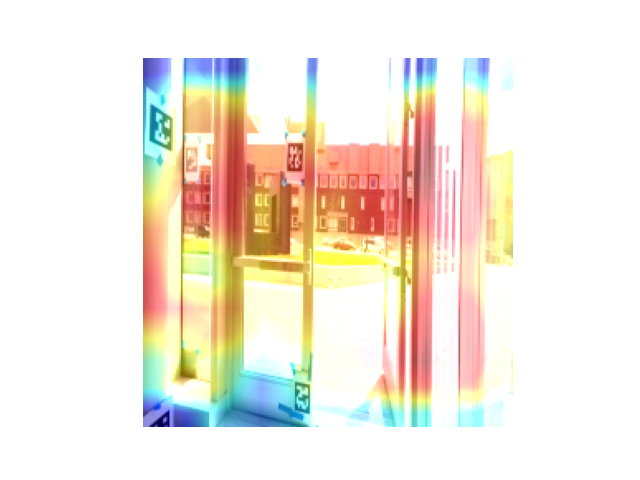}}
\subfloat[]{\includegraphics[clip, trim=3cm 1cm 3cm 2cm, scale=0.28]{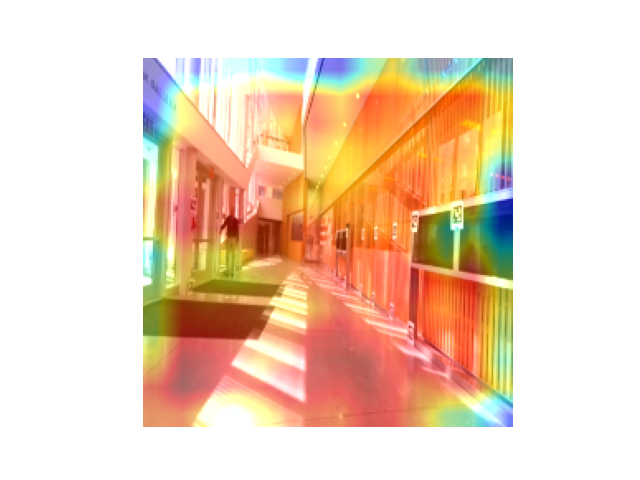}}

% \subfloat[]{\includegraphics[clip, trim=3cm 1cm 3cm 2cm, scale=0.28]{Figures/af_224_n.png}}
% \subfloat[]{\includegraphics[clip, trim=3cm 1cm 3cm 2cm, scale=0.28]{Figures/af_626_n.png}}
% \subfloat[]{\includegraphics[clip, trim=3cm 1cm 3cm 2cm, scale=0.28]{Figures/af_828_n.png}}

\caption{Activation maps over sample image triplet used for testing from PennCOSYVIO Dataset for bi-directional motion case, where maps in (a-c) are for query images, (d-f) are for corresponding (column wise) positive samples.}
% , and (g-i) are for the negative samples}
\label{act_penn_rev}
\end{figure}

To gain a better understanding of what the network has learned and what it sees in an image, we overlay the activation maps on their corresponding images for visual observation. The activation maps for the forward motion case are shown in Figures \ref{act_finn_for} and \ref{act_penn_for} for FinnForest and PennCOSVIO datasets, respectively. Similarly, the activation maps for the motion in the opposite direction (bi-directional case) are shown in Figures \ref{act_finn_rev} and \ref{act_penn_rev}.
For the forward motion case, it can be observed that almost the same regions are activated in the query and positive test image pairs, which is an intuitive conclusion. In contrast, the activation maps are flipped left-right in the backward motion case for the query image and the positive sample pair. This effect makes sense since the images are acquired from the opposite directions for roughly the same location.  This flip effect is dominant and easily observed from image pairs in Figure \ref{act_finn_rev}.a,d and \ref{act_finn_rev}.b,e. Moreover, the regions closer to the camera exhibit stronger activation compared to the regions that are far away. All these observations are in agreement with our hypothesis mentioned in Section \ref{data_Sel_SEC} that motivated the study.

\begin{figure*}[h!]
\centering
\includegraphics[clip, trim=0cm 0cm 0cm 0cm, width=1\textwidth]{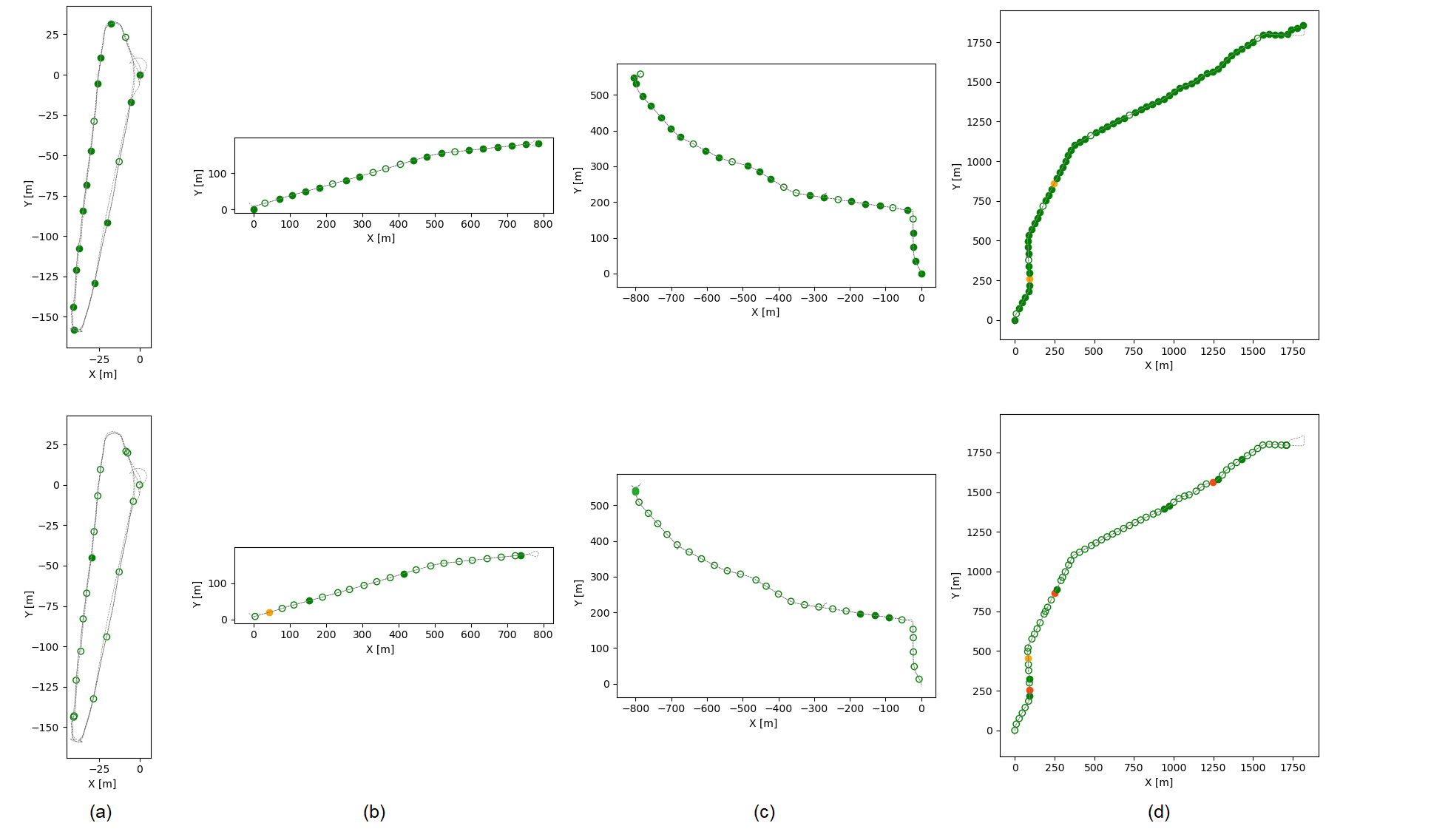}
\includegraphics[clip, trim=0cm 14.25cm 0cm 14.25cm, width=0.7\textwidth]{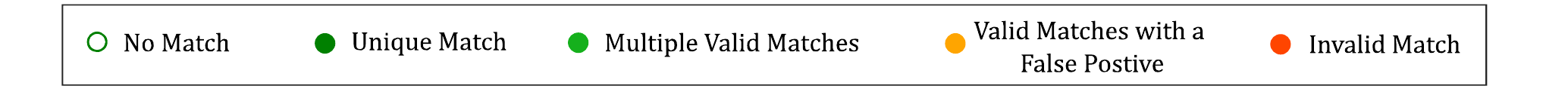}
\caption{ illustration of localisation results for FinnForest dataset. The top row shows the results for detections from the forward pass while the lower row shows the results for localisation from the backward pass.}
\label{finn_vlag_largecc}

\includegraphics[clip, trim=0cm 0cm 0cm 0cm, width=1.00\textwidth]{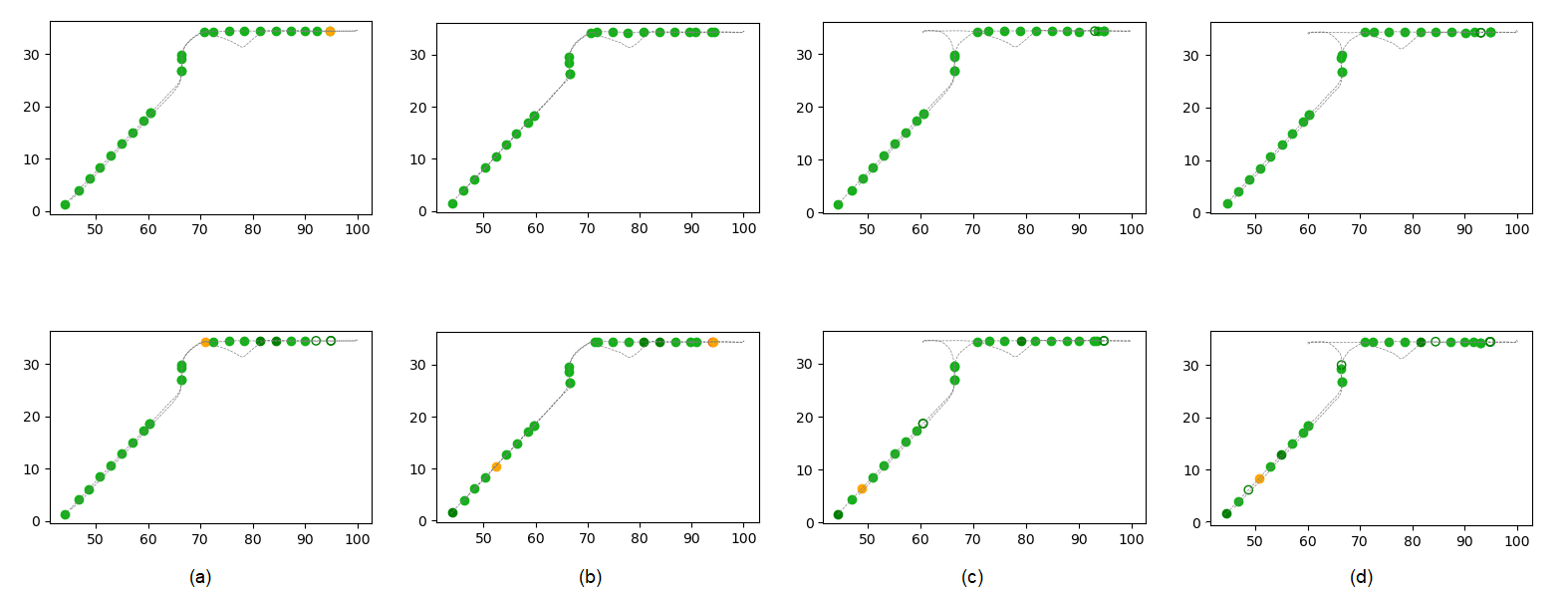}
\includegraphics[clip, trim=0cm 14.25cm 0cm 14.25cm, width=0.7\textwidth]{Figures/legend_loc.pdf}
\caption{ illustration of localisation results for PennCOSYVIO dataset. The top row shows the results for detections from the forward pass while the lower row shows the results for localisation from the backward pass.}
\label{penn_vlag_largecc}

\end{figure*}

PR curves are a good metric for binary classiﬁcation and to understand the overall performance of a system. However, to fully prove that our method is capable of accurately closing loops in practice, we perform loop closure offline in a causal manner. For this experiment, we simulate keyframe selection for loop closure at regular intervals based on the distance traveled. A loop closure candidate is detected, if the score of the query and a keyframe from the database is above a priori threshold $\tau$ and if it has passed the confidence sharing criterion among its neighboring localized keyframes. We eliminate the most recent images from the search space and wait until the database is large enough to start loop detection. The values of $\tau$ for the experiments shown in Figures \ref{finn_vlag_largecc} and \ref{penn_vlag_largecc} were selected from Figure \ref{PR_ Localisation} such that the recall rate is maximized with good precision returns. A slightly higher threshold was selected to illustrate all the possible outcomes in the experiments. As illustrated, a search can result in no match, a single unique match, multiple valid matches, a valid match with one or many false positives, or an invalid match. Multiple matches can be observed when the query images are acquired in open spaces and the scene does not change much among subsequent keyframes. In some cases, a keyframe can find a true and false positive at the same time due to visual similarity between multiple keyframes. This is more apparent in the FinnForest case where we observe quite frequent repetitive textures in the trees and on the road. Nonetheless, these false positives and invalid matches can be removed with the confidence-sharing approach between the neighboring keyframes.

It can be observed that the matches for the forward motion cases are significantly higher than the reverse case for the FinnForest. This makes sense since the forward motion provides more opportunities to observe similar scenes. In contrast, it is difficult to localize in the case of motion in the opposite direction for the FinnForest dataset as most of the scenes do not provide distinct landmarks within that short spatial window of observation. Nonetheless, the model was able to successfully identify the absence of matches and found enough matches that could be used to significantly improve odometry results. Even one valid match is enough to drastically improve odometry results and mitigate the errors due to drift that are accumulated in the odometry results. 

On the other hand, the model works very well on the PennCOSYVIO dataset since the indoor constrained environment provides distinct visual landmarks that are specific to their corresponding locations. Extrapolating on this observation, we can postulate that the approach would be effective in outdoor scene recorded within a city environment that provides distinct landmarks for bi-directional localisation. As mentioned earlier, we were not able to use the existing visual odometry datasets that were recorded in an urban environment as they are tailored for uni-directional odometry and SLAM purposes.

\subsection{Pose Regression}
 Earlier in Section \ref{poseReg_sec}, we discussed the pose regression block in detail and stressed the necessity of pose estimation through end-to-end learning. Here, we provide the experimental results of the proposed approach and compare the results against alternative approaches. The experimental results are tabulated in Tables \ref{res_pose_finn} and \ref{res_pose_penn} for FinnForest and PennCOSYVIO datasets, respectively. For comparison, we test the network by replacing the base model (VGG16) with Resnet50. Moreover, the relative impact of weight initialization is also studied by initializing the base models with weights acquired from models previously trained on ImageNet and Places 1365 (an extension of the Places 365 dataset) for classification tasks.
 
 For both datasets, the pose regression performance is measured as the absolute difference between the predicted and ground truth values for the location (in meters) and orientation (in degrees). Similar to localisation, we test the performance of the pose regression network and state the results on individual sequences and the combined case. The combined testing results are effectively the average of the individual results. The number of test image samples and the spatial extent of the area where each sequence was recorded are also mentioned.
 
 It can be observed that the network that has VGG16 as the base model and initialized with the weights of Places 1365 yields the best results followed by VGG16 initialized with ImageNet. This improvement was observed since Places 1365 incorporates visual scenes for scene classification that are quite relevant to our localisation task. In contrast, ImageNet is a more diverse dataset that is tailored for object classification. As a result, initialization with Places 1365 aids our network to generalize better to landscapes. On the other hand, Resnet50 performed poorly for both datasets. It is important to remember that the task at hand is localisation and not visual odometry. The pose regression is aimed at finding the relative pose between the query and a potential match for localisation. Traditional methods fail when we consider bi-directional cases of localisation. The results obtained for bi-directional pose regression in this study match the performance of other state-of-the-art approaches that are reported in studies conducted for uni-directional loop closure \cite{kendall2015posenet,laskar2017camera}.

\begin{table}[h]
\centering
\caption{Comparison of pose estimation results from the regressor model trained on FinnForest dataset. \label{res_pose_finn}}
\resizebox{\columnwidth}{!}{%
\begin{tabular}{|c|c|c|c|c|c|} 
\hline

\textbf{\makecell{Sequence }}&    \makecell{Test \\Samples} & \makecell{Spatial \\Extent (m)} & \makecell{Resnet50 \\Imagenet} & \makecell{VGG-Imagenet} & \makecell{VGG-Places 1365}  \\
                        \hline 
\textbf{S1}&     2044&      47 x 193  &          5.38m, 1.02$^{\circ}$ & 2.42m, 0.352$^{\circ}$ & 2.26m, 0.3$^{\circ}$ \\
                        \hline 
\textbf{S3}&     2706&      800 x 190 &          5.26m, 1.00$^{\circ}$ & 2.38m, 0.33$^{\circ}$ & 2.31m, 0.29$^{\circ}$  \\
                        \hline
\textbf{S4}&     3566&      812 x 568 &          5.68m, 1.06$^{\circ}$ & 2.54m, 0.39$^{\circ}$ & 2.36m, 0.32$^{\circ}$  \\
                        \hline
\textbf{S5}&     8866&      1826 x 1883 &        7.35m, 0.84$^{\circ}$ & 3.35m, 0.44$^{\circ}$ & 3.23m, 0.53$^{\circ}$  \\
                         \hline              
\textbf{Combined}&     17182&      2633 x 2014 &  5.92m, 0.98$^{\circ}$ & 2.67m, 0.38$^{\circ}$ & 2.54m, 0.36$^{\circ}$  \\
                        \hline
                        
\end{tabular}%
}
\end{table}

\begin{table}[h]
\centering
\caption{Comparison of pose estimation results from the regressor model trained on PennCOSYVIO dataset. \label{res_pose_penn}}
\resizebox{\columnwidth}{!}{%
\begin{tabular}{|c|c|c|c|c|c|} 
\hline

\textbf{\makecell{Sequence }}&    \makecell{Test \\Samples} & \makecell{Spatial \\Extent (m)} & \makecell{Resnet50 \\Imagenet} & \makecell{VGG-Imagenet} & \makecell{VGG-Places 1365}  \\
                        \hline 
\textbf{C2-af}&     3361&      144 x 36  &  3.79m, 0.71$^{\circ}$ & 1.51m, 0.21$^{\circ}$ & 1.35m, 0.22$^{\circ}$ \\
                        \hline 
\textbf{C2-bs}&     3330&      144 x 36 &       3.85m, 0.72$^{\circ}$ & 1.51m, 0.20$^{\circ}$ & 1.33m, 0.21$^{\circ}$ \\
                        \hline
\textbf{C2-bf}&     3090&      144 x 36 &       3.81m, 0.73$^{\circ}$ & 1.49m, 0.19$^{\circ}$ & 1.36m, 0.22$^{\circ}$  \\
                        \hline
\textbf{C2-bs}&     3375&      144 x 36 &       5.75m, 0.80$^{\circ}$ & 2.20m, 0.40$^{\circ}$ & 1.81m, 0.26$^{\circ}$  \\
                         \hline              
\textbf{Combined}&     13156&      144 x 36 &   4.3m, 0.74$^{\circ}$ & 1.68m, 0.25$^{\circ}$ & 1.46m, 0.22$^{\circ}$  \\
                        \hline
                        
\end{tabular}%
}
\end{table}

\section{Conclusion}
The article presents for the first time, to our knowledge, a learning-based approach to solve the problem of bi-directional loop closure in monocular images. We segregate the tasks of localisation into place identification and pose regression and solve them in two end-to-end deep learning steps. We demonstrate that it is indeed possible to achieve bi-directional loop closure on monocular images by carefully posing the problem and leveraging the training data for the networks. Moreover, we demonstrate that the networks generalize well and aim for learning the geometric and spatial relations in images rather than memorize the scenes/locations. This is validated by the performance of the model on unseen data.
We compare the proposed approach with other deep learning methods and classical approaches and demonstrate superior performance for localisation. Moreover, we provide both qualitative and quantitative results to corroborate the claim.
A natural extension of the work would be to extend the case scenarios and test the approach with more datasets.

%=========================== BIBLIOGRAPHY============================
\bibliographystyle{IEEEtran}
\bibliography{mainD.bib}

% \newpage

\section{Biography Section}

\vspace{-33pt}
\begin{IEEEbiography}[{\includegraphics[width=1in,height=1.25in,clip,keepaspectratio]{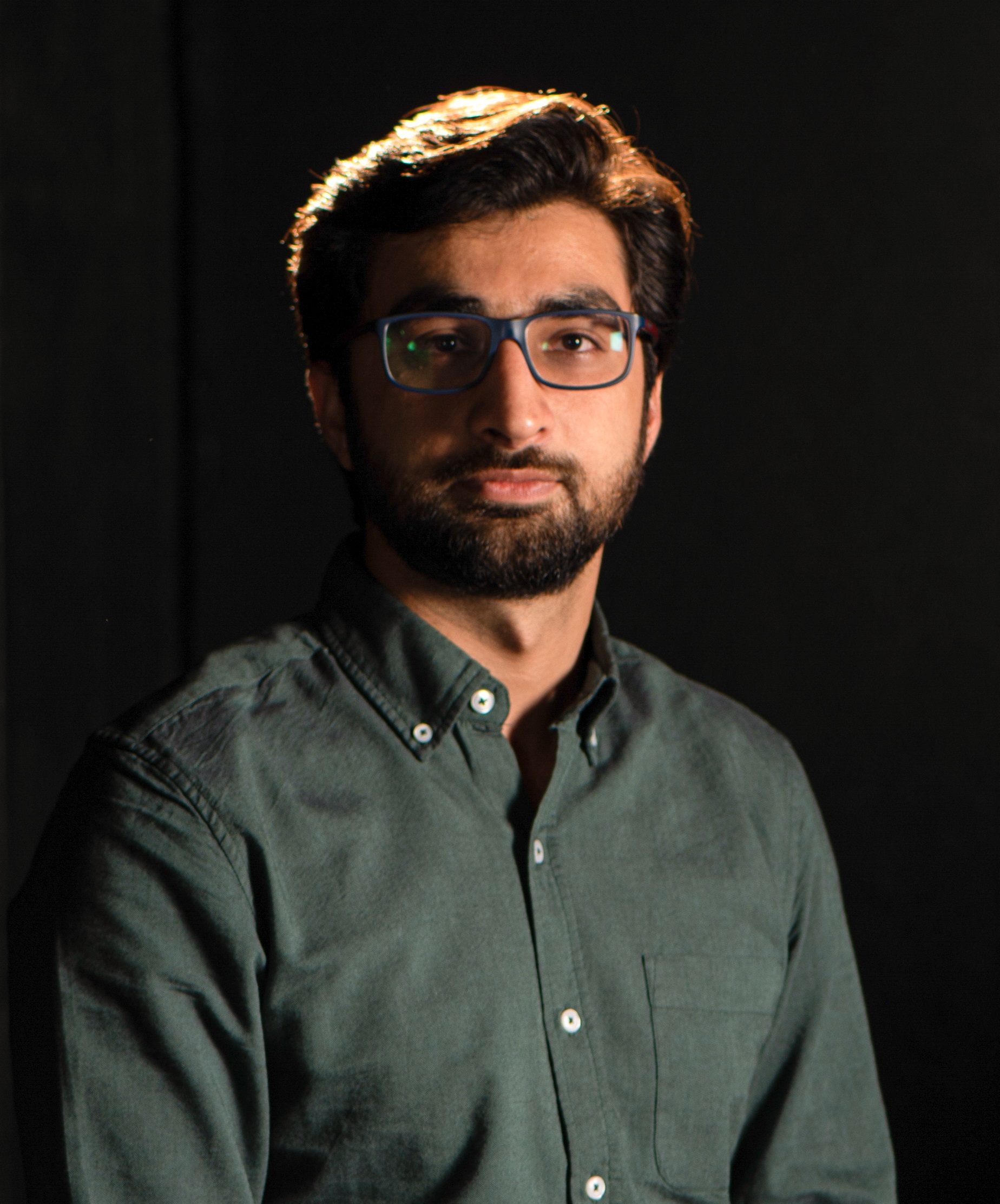}}]{Ihtisham Ali}
received his B.Sc. in Mechatronics Engineering from the University of Engineering and Technology, Pakistan (2014) and his M.Sc. in Automation Engineering from Tampere University, Finland (2017). Currently, he is a doctoral researcher in 3D Media Group at Tampere University. He has worked on several industrial projects pertaining to machine automation using visual cues. His research interest is focused on computer vision and robotics specifically object pose estimation, 3D reconstruction, and visual SLAM.
\end{IEEEbiography}

\vspace{11pt}

\vspace{-33pt}
\begin{IEEEbiography}[{\includegraphics[width=1in,height=1.25in,clip,keepaspectratio]{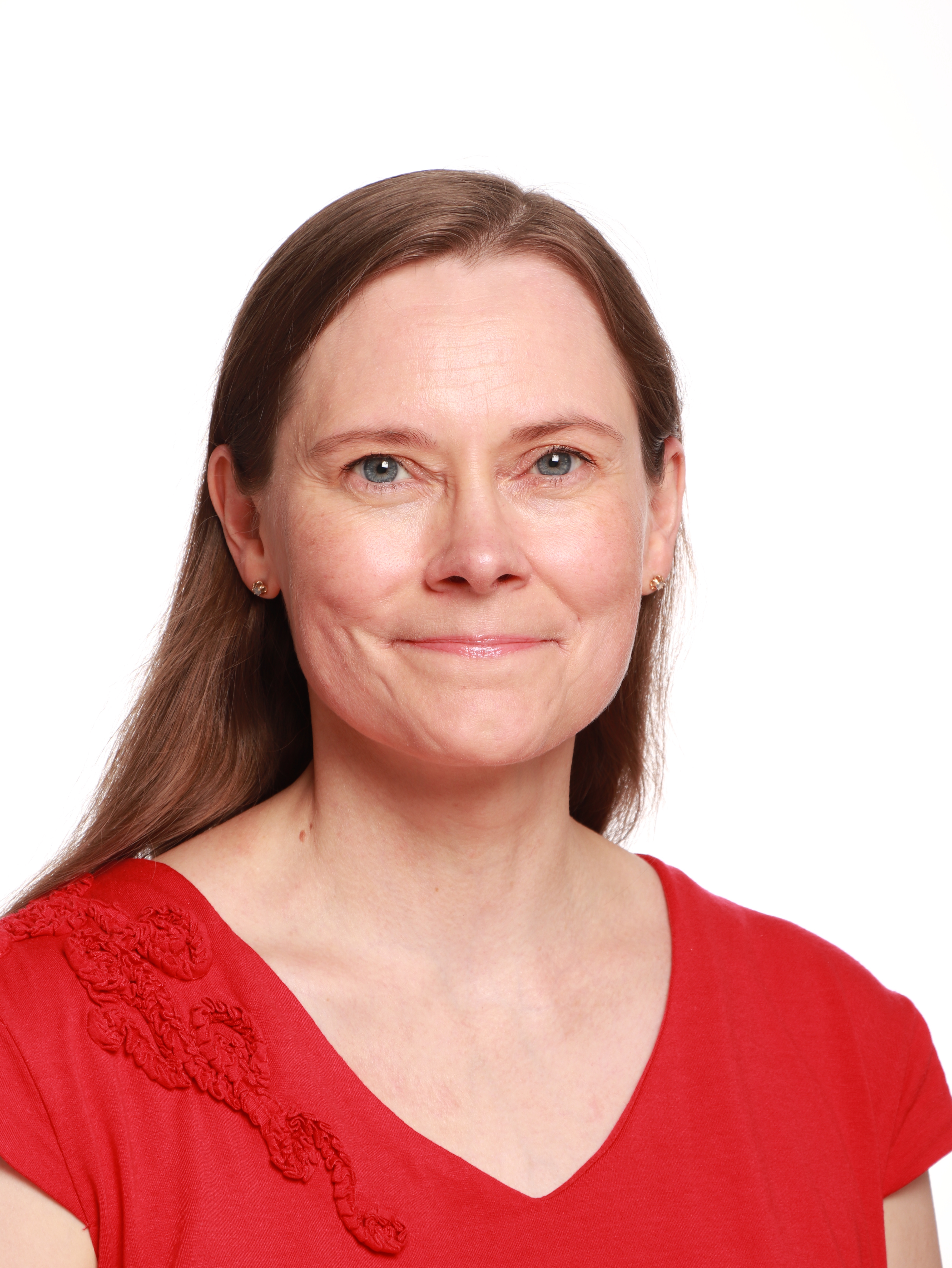}}]{Sari Peltonen}
received the M.Sc. degree in mathematics from the University of Tampere, Finland, in 1996. She received the Ph.D. degree in signal processing from the Tampere University of Technology in 2000. She is currently University Lecturer in signal processing at the Unit of Computing Sciences in Tampere University. Her research interests include robust estimation, image processing and tomographic image reconstruction.
\end{IEEEbiography}

\vspace{11pt}

\vspace{-33pt}
\begin{IEEEbiography}[{\includegraphics[width=1in,height=1.25in,clip,keepaspectratio]{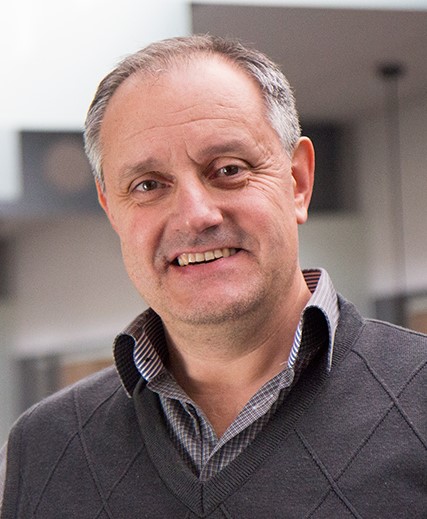}}]{Atanas Gotchev}
received his M.Sc. degrees in radio and television engineering (1990) and applied mathematics (1992), his Ph.D. degree in telecommunications (1996) from the Technical University of Sofia, and the D.Sc.(Tech.) degree in information technologies from the Tampere University of Technology (2003). He is a Professor of Signal Processing and Director of the Centre for Immersive Visual Technologies at Tampere University. His recent work concentrates on the algorithms for multi-sensor 3-D scene capture, transform-domain light-field reconstruction, and Fourier analysis of 3-D displays.
\end{IEEEbiography}

\vfill

\end{document}